\documentclass{article}

\usepackage[preprint]{neurips_2019}

\usepackage[utf8]{inputenc} 
\usepackage[T1]{fontenc}    
\usepackage{hyperref}       
\usepackage{url}            
\usepackage{booktabs}       
\usepackage{amsfonts}       
\usepackage{nicefrac}       
\usepackage{microtype}      
\usepackage{graphicx}
\usepackage[fleqn]{amsmath}
\usepackage{setspace}
\usepackage{amssymb}
\usepackage{wrapfig}
\title{Inference with Hybrid Bio-hardware Neural Networks}

\author{%
  Yuan Zeng $^*$ \quad Zubayer Ibne Ferdous$^*$  \quad Weixiang Zhang$^\dag$ \\
  \textbf{Mufan Xu}$^\dag$ \quad \textbf{Anlan Yu} $^*$ \quad \textbf{Drew Patel} $^*$\\
  \textbf{Xiaochen Guo}$^*$ \quad  \textbf{Yevgeny Berdichevsky}$^*$$^\S$\quad \textbf{Zhiyuan Yan}$^*$   \\\\
  Lehigh University, $^*$Electrical and Computer Engineering Department $^\S$Bioengineering Department \\
  $^\dag$Beihang University, Electrical and Computer Engineering Department \\
}

\begin{document}

\maketitle

\begin{abstract}
To understand the learning process in brains, biologically plausible algorithms have been explored by modeling the detailed neuron properties and dynamics. 
On the other hand, simplified multi-layer models of neural networks have shown great success on computational tasks such as image classification and speech recognition. However, the computational models that can achieve good accuracy for these learning applications are very different from the bio-plausible models. 
This paper studies whether a bio-plausible model of a \textit{in vitro} living neural network can be used to perform machine learning tasks and achieve good inference accuracy. 
A novel two-layer bio-hardware hybrid neural network is proposed. 
The biological layer faithfully models variations of synapses, neurons, and network sparsity in \textit{in vitro} living neural networks.
The hardware layer is a computational fully-connected layer that tunes parameters to optimize for accuracy. 
Several techniques are proposed to improve the inference accuracy of the proposed hybrid neural network. For instance, an adaptive pre-processing technique helps the proposed neural network to achieve good learning accuracy for different living neural network sparsity. The proposed hybrid neural network with realistic neuron parameters and variations achieves a 98.3\% testing accuracy for the handwritten digit recognition task on the full MNIST dataset.
\end{abstract}

\section{Introduction}
\vspace{-0.5em}
\label{introduction}
Artificial neural networks (ANNs), especially with deep network structures, have achieved great success on machine learning applications such as image classification~\cite{vision} and speech recognition~\cite{speech}. 
Prior works have proposed techniques to improve network accuracy while reducing training time and improving computational efficiency, for example, network quantization~\cite{quantization}, binarization~\cite{binary}~\cite{xnor}, compression~\cite{compression}, and pooling~\cite{pooling}. 
ANNs rely on static and numerical abstractions of the brain function. However, the brain is a dynamical system that processes information through spikes. Spiking neural networks (SNNs)~\cite{snn} can model the biological neurons and synapses at different levels of details~\cite{dendritic1}~\cite{dendritic2}~\cite{inhibitory1}~\cite{inhibitory2}~\cite{recurrent}. For example, the leaky integrate and fire (LIF) model~\cite{inf} can capture the "integration" process of the synaptic current that triggers the action potential, which is widely used for neuromorphic computing that targets low power consumption~\cite{neuromorphic}. 
However, those works focus on one or several specific neuron properties and do not captures all of them. Artificial and spiking neural networks have been inspired by studies of the brain's circuitry.  However, brain's neurons are substantially more complex than their simplified models used in ANN and SNN, and it is not clear whether networks of biological neurons actually perform learning and computational tasks in the same way as ANNs or SNNs.
 
 This paper explores the learning capability of the \textit{in vitro} living neural network, which can be generated by dissociating a rodent cortex into individual cells, placing them on an adhesive dish, and maintaining them in physiological conditions for several weeks~\cite{neuron_circuit1}. Neurons in a dish make random synaptic connections with each other, forming an \textit{in vitro} living network. These networks have been used to study mechanisms of biological computation and learning ~\cite{stdp1}~\cite{neuron_circuit4}~\cite{neuron_circuit5} and construct simple logic gates~\cite{neuron_circuit2}. The idea of using living neural networks to perform machine learning tasks have been proposed in a prior work~\cite{icassp}. The feasibility was established and a 74.4\% accuracy was reported for the handwritten digit recognition task. A "Learning machine" composed of living neurons can be used as a platform to validate computational algorithms in biological experiments, and may be more energy-efficient as compared to the silicon-based neuronal network implementations~\cite{energy}.

In this paper, an inference mechanism targeting the \textit{in vitro} living neural network implementation is introduced and its performance on the MNIST dataset is computationally assessed. The algorithm is validated in a biophysical model of a living neural network with experimentally derived parameters that capture realistic neural, synaptic, and connectivity properties and variability. This work aims to demonstrate a bio-plausible inference mechanism that: (1) achieves high classification accuracy and (2) can be implemented using a living neural network. 

\begin{figure}
  \centering
  \includegraphics[width=13.9cm,height=4.5cm]{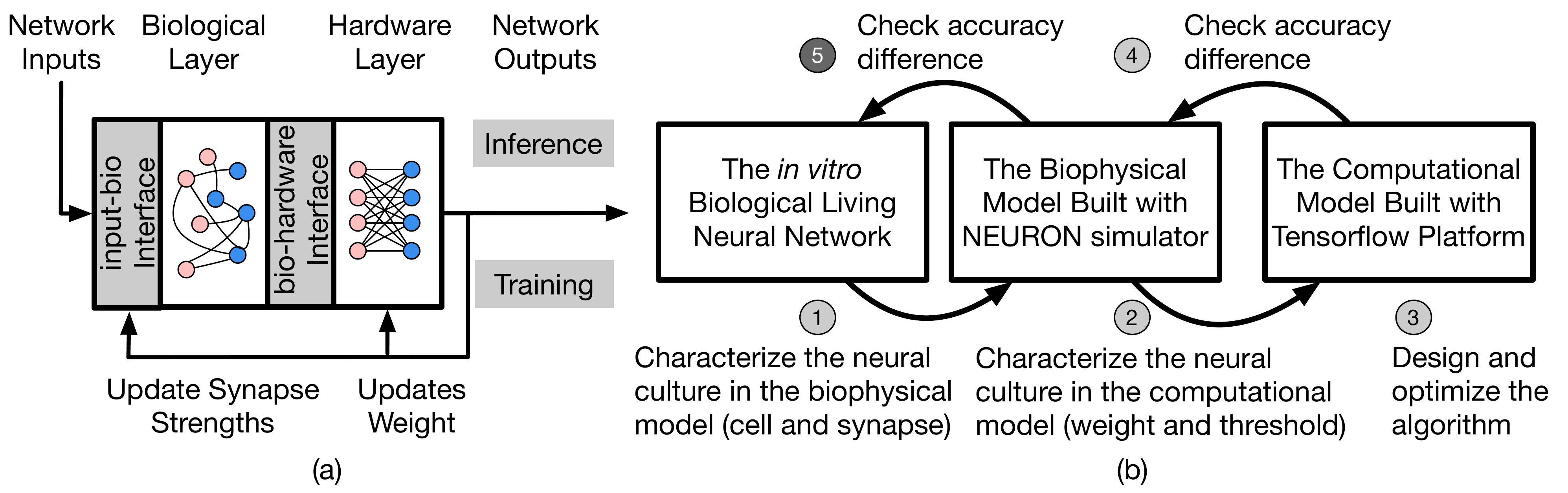}
  \vspace{-2em}
  \caption{Key idea of the paper: (a) Two-layer hybrid bio-hardware neural network (b) A three step method for algorithm design with living neural networks.}
  \label{keyidea}
  \vspace{-1.5em}
\end{figure}

A novel two-layer hybrid bio-hardware neural network is proposed (Fig~\ref{keyidea} (a)). The first layer is to be implemented in an \textit{in vitro} living neural network. The second layer is computational and can run on general propose computers or accelerators. Inputs to the living neural network are applied through an input-bio interface to the input (pink cells) in the biological layer. The network activity is measured from the living neural network outputs (blue cells) and converted to the digital representation through the bio-hardware interface. 
These interfaces are not restricted to a specific technology, and may be implemented via optogenetics ~\cite{optical1}~\cite{optical2} or high-density microelectrode arrays (MEA)~\cite{MEA1}~\cite{MEA2}~\cite{MEA3}. In order to faithfully model a living neural network, the biological layer uses realistic parameters derived from experimental characterization of rat cortical \textit{in vitro} networks.
The hardware layer provides both high-precision data representation and flexibility for weight updates, and hence has the potential to boost the performance. 

\subsection{Key Contributions}
\vspace{-0.5em}
In this paper, a two-layer bio-hardware hybrid neural network is proposed for inference and its performance on the handwritten digit recognition task is evaluated. A supervised back-propagation learning algorithm (Section~\ref{network}) is used to calculate weight and synaptic strength updates under constraints derived from experimental characterization (Section~\ref{method}).   

For algorithm development, a three-stage method involving a living neural network, a biophysical model, and a computational model (Section~\ref{method}) is used. The biophysical model captures the living neural network properties and variability by neuron and synapse parameter fitting from experimental data. However, the simulation speed for the biophysical layer is slow because of the large number of differential equations that are involved in modeling ion channels and synapses. Therefore, a simplified computational model with fast simulation speed is used and the living neural network properties are transferred into this model by fitting the weight and threshold distribution (Section~\ref{culture_transfer}).  

To improve the inference performance of the proposed hybrid neural network, several techniques are proposed (Section~\ref{optimization}). An adaptive pre-processing technique helps the the hybrid network to achieves good performance despite the living neural network limitations and variability. Gradient estimator optimization approach, computational layer parameter tuning, and adaptive learning rate are also explored. 

\subsection{Living Neural Network Properties and the Difference from Prior Work}
\vspace{-0.5em}
This is the first work to develop algorithms tailored for hybrid bio-hardware neural networks. The \textit{in vitro} living neural networks have unique properties and constrains: 
(1) A synaptic weight cannot change between positive (excitatory neuron) and negative (inhibitory neuron)~\cite{inhibit}. 
(2) Synaptic weight changes are constrained to a range of 2$\times$ larger and 0.5$\times$ smaller of the initial weights~\cite{stdp1}. 
(3) The network has inherent variability in neural, synaptic, and connectivity parameters~\cite{scaling}. 
(4) The neurons are sparsely connected and the connectivity is reported to be less than 40\%~\cite{scaling}. 
(5) The connections in an \textit{in vitro} network are randomly formed. For a network with pre-defined inputs and outputs, not only input-output connections are established, but input-input, output-output, and output-input connections also exist~\cite{recurrent}. 

Although some of the existing algorithm studies used networks characterized by one or more properties listed above, none of them used the full constraints inherent to a living neural network.  The work in~\cite{icassp} proposed a supervised learning algorithm for \textit{in vitro} living neural networks.
However, the work in ~\cite{icassp} does not consider random and sparse connections.
Compared to the 74.4\% accuracy on 10000 MNIST images reported by~\cite{icassp}, the best result achieved by this paper is 98.3\%.  

\section{Experimental Assumptions and the Algorithm Design Method}
\vspace{-0.5em}
\label{method}
To perform the learning task, this work uses binary data representation.
Living neural network is randomly connected, which means other than the input-output connections, connections also exist among inputs, among outputs, and from outputs to inputs. 
As a result, the network has poly-synaptic (secondary) spikes triggered indirectly by the inputs, in addition to single-synaptic (primary) spikes. 
To avoid this issue, the proposed work uses a deterministic input instead of spike trains that are used in rate and temporal coding.  
To distinguish the primary spikes from the secondary spikes, this paper assumes an early "cut-off" mechanism can be applied in the experiments. This is feasible because most of the primary spikes happen before the secondary spikes as shown in the supplementary materials. 

To efficiently design the algorithm for living neural networks, a three-stage method is proposed as shown in Fig~\ref{keyidea} (b). The first stage is the biological platform which contains the \textit{in vitro} living neural networks; 
the second one is a biophysical model which uses the NEURON simulator~\cite{NEURON} to build two-compartment Pinsky-Rinzel model~\cite{pr2} for neuron, and alpha model~\cite{alpha} for synapse; 
and the third one is a computational model built using the tensorflow~\cite{tensorflow} to model the hard threshold behavior of neurons. 
The purpose of using the three-stage method is to improve the speed of algorithm exploration without losing fidelity.
The study consists of the following steps:
1) biological experiments are conducted to determine neural and synaptic parameters; 
2) the biophysical model of the \textit{in vitro} neural network is created using experimentally obtained parameters. Variability in the biophysical model is then quantified and transferred into weight and threshold variations in the computational model;
3) learning algorithm is designed and optimizations are applied in the computational model with fitted biological layer parameters; 
4) accuracy of the algorithm is checked on the biophysical model; 
5) accuracy of the developed algorithm is checked on the living neural network. Design details of step 1-4 are introduced in the rest of the paper and step 5 remains as future work. Through this method, the living neural network properties can be converted into a simple computational model for fast algorithm design and optimization. 

\section{Network Topology and the Learning Algorithm}
\vspace{-0.5em}
\label{network}

\begin{figure}
  \centering
  \includegraphics[width=13.8cm,height=8cm]{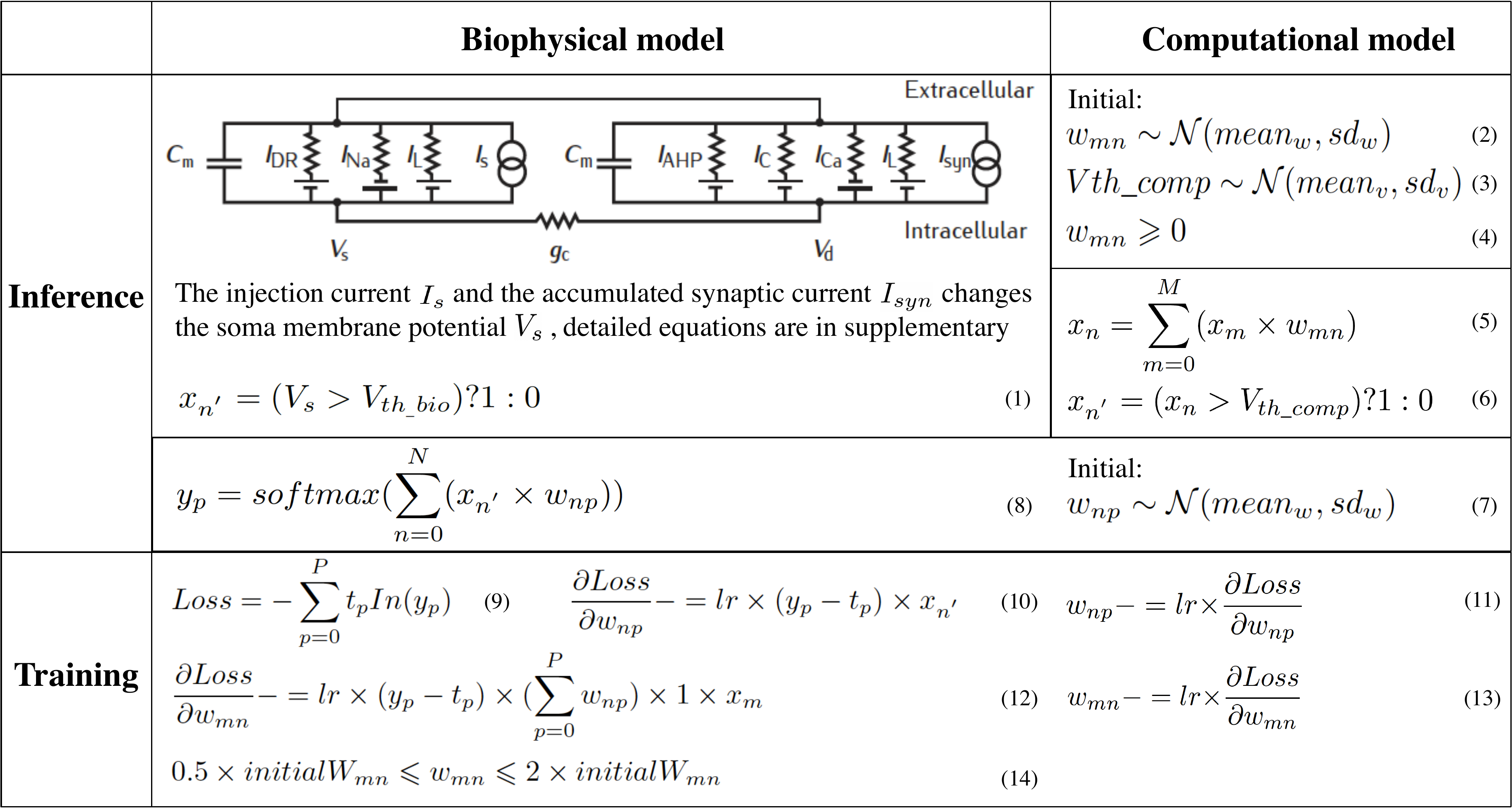}
  \vspace{-0.5em}
  \caption{The hybrid network learning algorithm~\cite{pr}~\cite{alpha}~\cite{cross}.}
  \label{algorithms}
  \vspace{-1.5em}
\end{figure}

\begin{wrapfigure}{r}{0.44\textwidth}
  \begin{center}
  \vspace{-0.8em}
    \includegraphics[height=4.2cm]{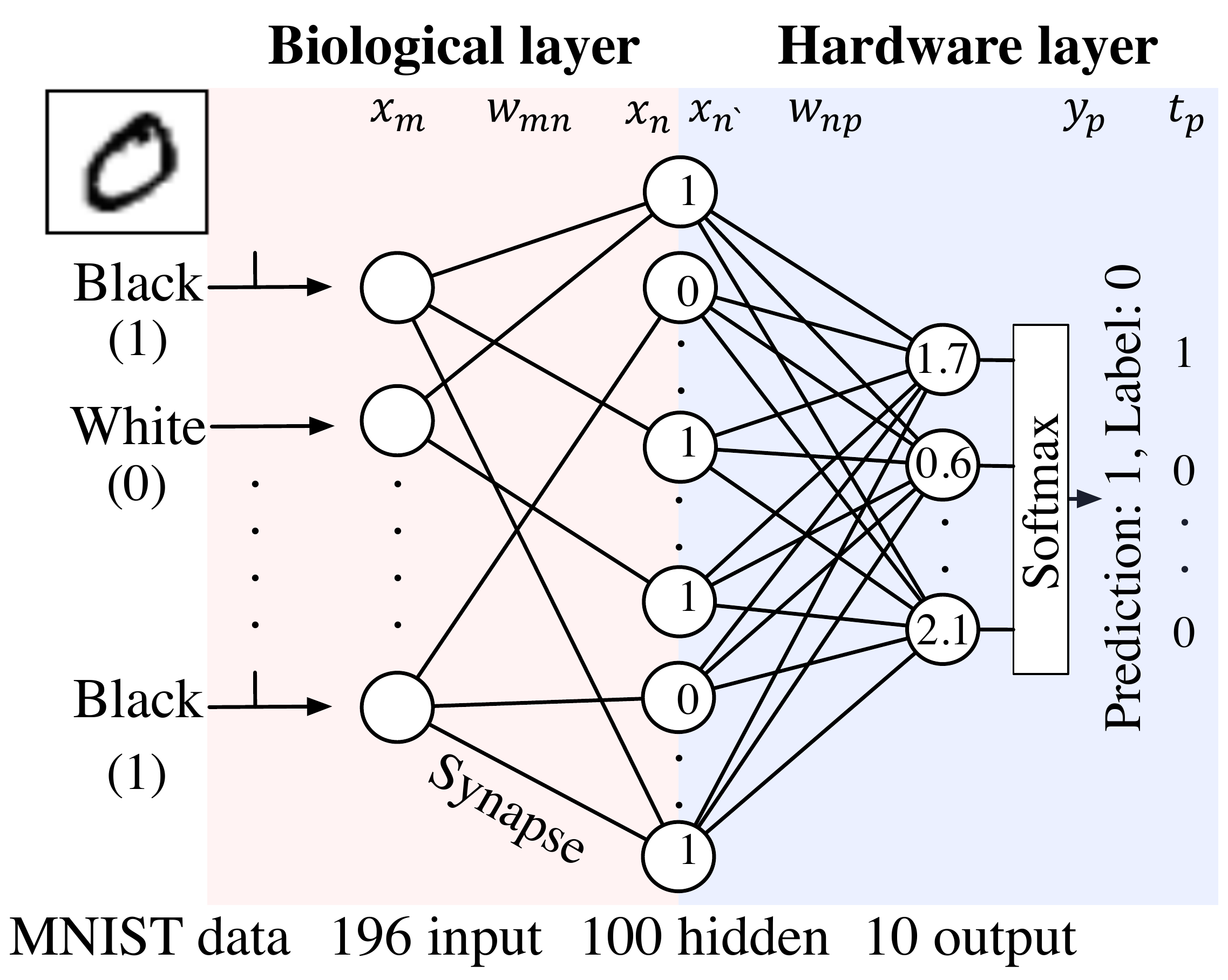}
  \end{center}
  \vspace{-1em}
  \caption{An example of using the hybrid neural network for digit recognition, the hidden layer is the output of the biological layer and the input of the computational layer.}
  \vspace{-1em}
  \label{network_example}
\end{wrapfigure}

This work uses the MNIST dataset to evaluate the performance of the proposed hybrid bio-hardware neural network. 
Each MNIST image has $28\times28$ pixels in gray-scale. However, controlling 784 input neurons is difficult, and each MNIST image is compressed to $14\times14$ pixels. Section 5.2 describes the details of different pre-processing methods to compress the images.
After the pre-processing and compression, a threshold is set to 100 to turn the pixel value ranges between 0-255 into binary representation. More details of the image pre-processing are discussed in Section~\ref{optimization}. 
The \textit{in vitro} neurons are divided into input and output neurons. Each input neuron corresponds to a pixel in the image. If the pixel value is one (Fig~\ref{network_example}), the corresponding neuron is stimulated optically or electrically to generate a spike. 
The stimuli can be given to different input neurons at the same time. 


The biological layer with a 40\% connectivity is modeled (Fig~\ref{network_example}). To capture the neuron dynamic in the biophysical model, a two-compartment Pinsky-Rinzel neuron model~\cite{pr2} with three somatic ion channels and four dendritic ion channels is used (Fig~\ref{algorithms}). An alpha synapse model~\cite{alpha} is used. 

The model details are provided in the supplementary materials. All of the neurons are modeled as excitatory in the proposed work. 
The injection current ($I_s$) for pre-synaptic neurons is used as input to the network. For post-synaptic neurons, an action potential is triggered when the accumulated synapse current ($I_{syn}$) is large enough. 
In the biophysical model, a spike is considered to occur when the somatic voltage ($V_s$) exceeds zero (Eq. (1)), which happens near the peak of the action potential. For the computational model, weights and the neuron thresholds for the biological layer are initialized following a normal distribution (Eqs. (2) and (3)), and no negative weight is allowed (Eq. (4)). If the accumulated inputs-weights product at a certain neuron exceeds the threshold (Eqs. (5) and (6)), a spike is generated.

The hardware layer is fully connected with ten outputs (Fig~\ref{network_example}), the weights are initialed following a normal distribution without any constraint (Eqs. (7)). A softmax function~\cite{cross} is used to normalize the output (Eqs. (8)). The index of the largest output is the prediction result. The cross entropy loss~\cite{cross} is used for error backpropagation (Eqs. (9)). The gradient of the nondifferentiable hard threshold function is estimated as constant one, which is known as the "straight through estimator~\cite{estimator}" (Eqs. (10)-(13)). The weights of the biological layer are restricted to the range of 0.5$\times$--2$\times$ of initial weights (Eqs. (14)).

\section{Characterization of the Living Neural Networks}
\vspace{-0.5em}
\label{culture_transfer}

\subsection{Model Neural Variations in the Biophysical and the Computational Models}
\vspace{-0.5em}

To capture the living neural network variations in the biophysical model, parameters of the neuron and synapse models are fitted to data obtained through intracellular recordings of nine different cells and 12 different synapses. 
The parameter fitting results are reported in the supplementary materials. 

To reduce the computational complexity, this work further converts the biophysical model into a computational neural network model with a hard threshold function. To ensure that a similar accuracy can be achieved after the conversion, this paper uses the minimum number of pre-synaptic neurons that triggers a post-synaptic neuron to fire (minPreNum) as the bridge to convert the variations in the biophysical model to the variations in the computational model.  

\begin{figure}
  \centering
  \includegraphics[width=12cm]{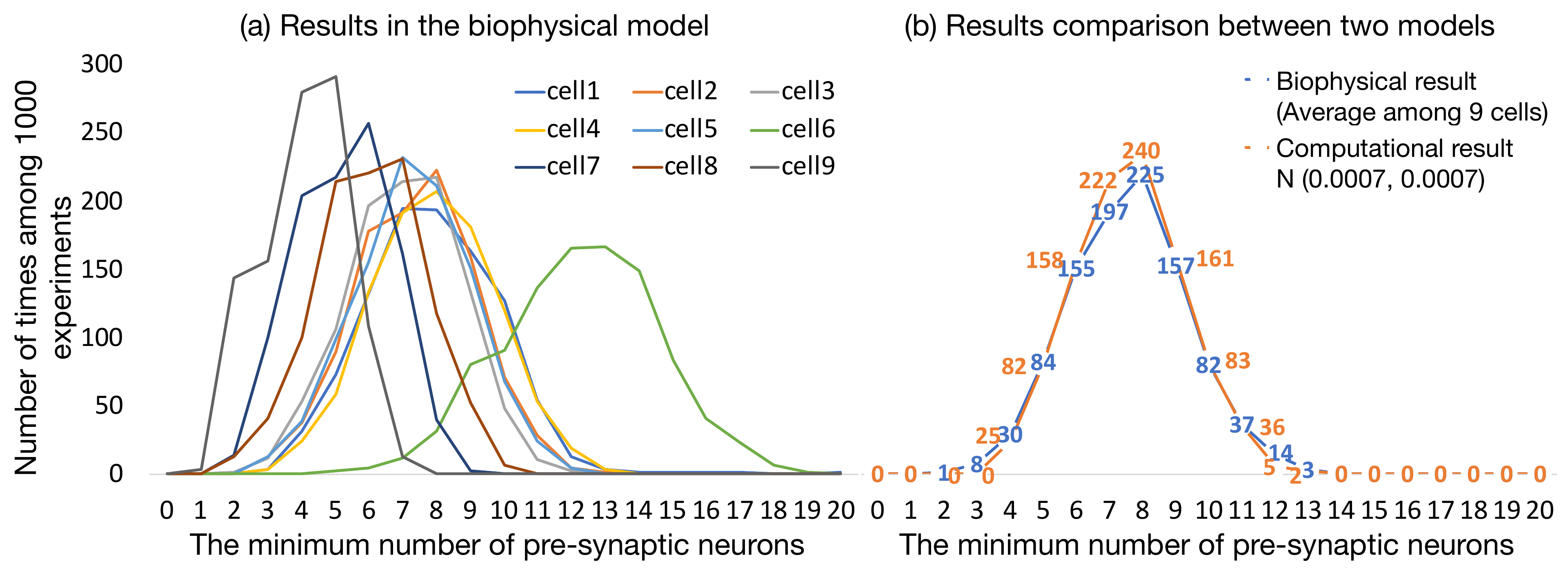}
  \vspace{-0.5em}
  \caption{Converting biophysical variations into the threshold and weight variations.}
  \label{variation1}
  \vspace{-1.5em}
\end{figure}

 To convert variability in the biophysical parameters to minPreNum, the following experiment is conducted in the biophysical model.  The number of pre-synaptic neurons is varied from 1 to 20 and the pre-synaptic neurons are stimulated through injected current ($I_s$). 
The post-synaptic neuron is sequentially selected from the fitted neuron 1-9. The input neurons and synapses are randomly selected from the nine fitted neurons and 12 fitted synapses respectively. A selection is allowed to repeat. The experiment is repeated 1000 times in simulation and the results are shown in Fig~\ref{variation1} (a). Different post-synaptic neurons have different minPreNum curves. 

The threshold and weight variations are assumed to follow a normal distribution. The expectations of each of the minPreNum curves are used to estimate the threshold variation. The average of the nine minPreNum expectations is 7.2, and the standard deviation is 2.1. These values are multiplied with the average maximum synapse conductance (the average gsyn 0.0008 in the supplementary materials is used to estimate the average weights) to derive the threshold distribution N(0.0058, 0.0017).
To estimate the weight distribution, the nine minPreNum are aligned with peak and the aligned points on the curves are averaged into one curve in Fig~\ref{variation1} (b), which is used to calculate the weight distribution. To fit the mean and standard deviation of the curve, the minPreNum experiment is repeated using the computational model. A post-synaptic neuron with threshold 0.0058 is used. A weight distribution of N(0.0007, 0.0007) has the best fit.

\subsection{Comparing the Biophysical and the Computational Models}
\begin{figure}
  \centering
  \vspace{-1.5em}
  \includegraphics[width=13.8cm,height=5cm]{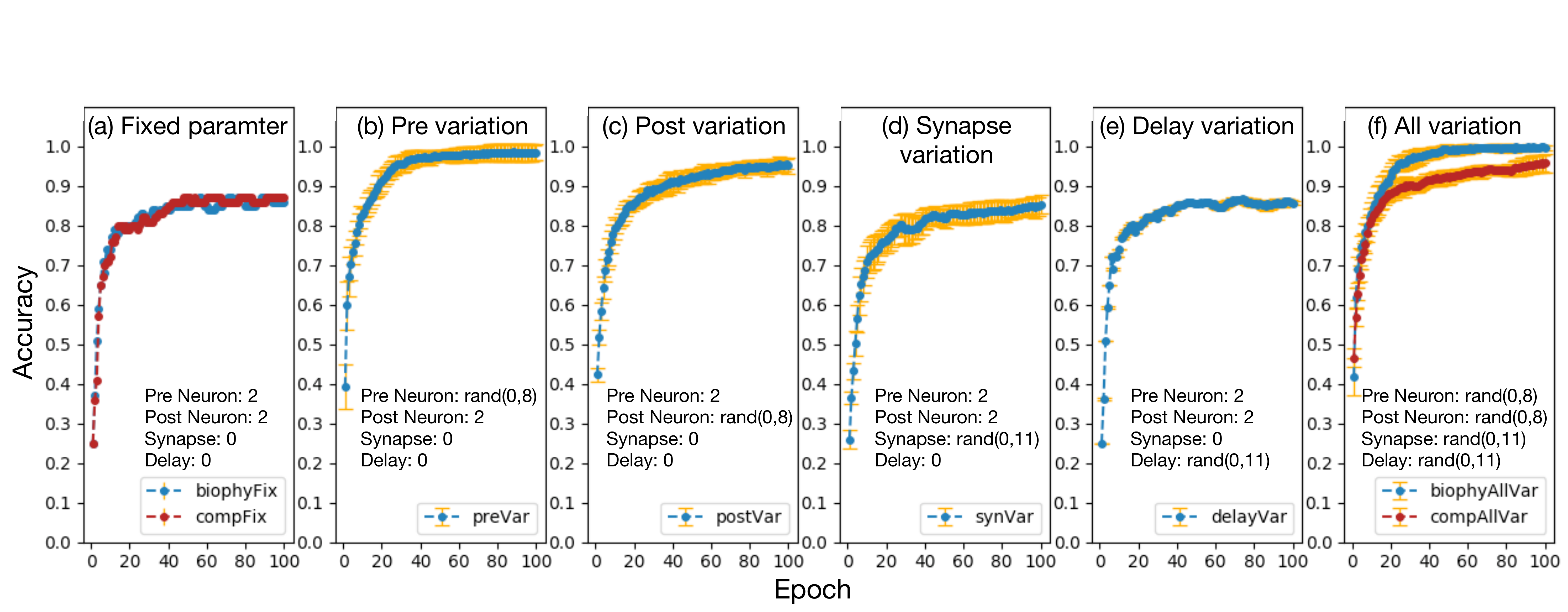}
  \vspace{-1em}
  \caption{Variation study of the computational and the biophysical models.}
  \label{variation2}
\end{figure}

\vspace{-0.5em}
To validate the computational model against the biophysical model, the handwritten digit recognition task is performed with both the computational and biophysical models. Experiment settings are listed in Table~\ref{100setting} and the testing accuracies are compared in Fig~\ref{variation2}. For each experiment, both the mean accuracy and the standard deviation among ten trials are reported. As a first step, the computational and biophysical models are compared without any variations. In this experiment, a fitted neuron (neuron 3) and a fitted synapse (synapse 1) are chosen from Table~\ref{fitting_result} in the supplementary materials.  The weights in the computational model are initialized to the gsyn value of the fitted synapse 0 without variation. Vth in the computational model is set to 0.0055 without variation to match the selected biophysical model. The network connectivity, topology, and the hardware layer weights are initialized to exactly the same for the biophysical and computational models. Fig~\ref{variation2} (a) shows that the accuracy results are nearly the same for these two models. 

Figure~\ref{variation2} (b)-(e) present the variation influence of the pre-synaptic neurons, post-synaptic neurons, synapse(gsyn and $\tau$), and the transmission delay. The transmission delay is defined as the difference between the time of the peak of the pre-synaptic spike and the time when the post-synaptic membrane potential starts to raise.
The results show that the digit recognition accuracy is higher with the pre-synaptic and post-synaptic neuron variations. Synapse and transmission delay variations do not influence the accuracy too much. This is because neuron variations have a greater impact on the weight and threshold variations as compared to the synapses. However, synapse variations lead to more noise as suggested by the error bar. This is because synapse variations can change the dynamic behaviors of how synaptic current decays and hence influence whether the post-synaptic neuron can fire. Transmission delay does not have a large variation due to the limited space the living neuron culture reside in. For the final step, all of the variations are added to both the biophysical and the computational model (Fig~\ref{variation2} (f)). 
With the variations, the accuracies are higher for both models as compared to no variation, and the accuracy of the biophysical model is higher as compared to the computational model.
The reason behind this accuracy difference will be explained in Section~\ref{optimization}. 

\section{Optimization Method and Simulation Results}
\vspace{-0.5em}
\label{optimization}

\begin{table}
    \vspace{-0.5em}
    \caption{Simulation parameters for the variation study.}
    \vspace{-0.5em}
    \label{100setting}
    \footnotesize
    \centering
    \begin{tabular}{ll}
     \hline
     Platform        &Floating point 64 for computational layer  \\
     Dataset/Network &100 MNIST images, training set same as the testing set, size:196-100-10   \\ 
     Bio layer(fix) &Sparsity:40\%, $Vth_{cp}:N(0.0055, 0)$, $initW_{cp}:N(0.00072, 0)$, Lr:1e-4  \\
     Bio layer(var) &Sparsity:40\%, $Vth_{cp}:N(0.0058, 0.0017)$, $initW_{cp}:N(0.0007, 0.0007)$, Lr:1e-4 \\
     Hardware layer  &Fully connected, $initW$:N(0.0007, 0.0007), Lr:1e-2 \\
     \hline
\end{tabular}
\vspace{-1em}
\end{table}

In this section, the fitted computational model is used for algorithm optimization. The reason behind the accuracy gap between the hybrid network and other similar neural network is analyzed. Three optimization methods are explored to improve the accuracy using 1000 MNIST images. The performance of the proposed hybrid neural network is evaluated using the full MNIST dataset (60000 training images, 10000 testing images).

\subsection{The Average of Firing Neurons in the Hidden Layer}
\vspace{-0.5em}
When using the hybrid network to perform the learning task based on the algorithm shown in Figure~\ref{algorithms}, the accuracy (85.5\%) after parameter tuning is lower than a 196-100-10 binary network~\cite{binary} tested on the same dataset (100\% at epoch 100). 
The hidden layer of the binary network applies a hard threshold function and the weights are binaries. The success of the binary network~\cite{binary} suggests that the proposed hybrid network should have similar learning capability. After a closer examination of the hybrid network, one hypothesis is that the average number of firing neurons in the hidden layer ($Nf_{hidden}$) during the learning process directly influences the network capacity and hence the accuracy. When half of the hidden layer neurons spike on average, the network has the best learning capability.
An intuitive example is that, if none of the neurons in the hidden layer spike, no matter what images are given from the dataset, the network will not learn at all. Similar situation also happens if all of the neurons in the hidden layer spike regardless of the input image. 

\setcounter{equation}{14}
Considering the network structure in Fig~\ref{network_example}, we introduce a parameter that is related to $Nf_{hidden}$:
\begin{equation}
f = (Sparsity \times Nin_b \times meanW)/Vth,    
\end{equation}
where $Sparsity$ is calculated by the number of connected neuron pairs divided by the total number of neuron pairs between the input- and the hidden- layer neurons,
$Nin_b$ represents the number of black pixels in the input images, 
$meanW$ represents the average weights between input- and hidden- layer neurons for the entire dataset after training, network learning rate can change the $meanW$ value, and $Vth$ is the average threshold of the hidden layer neurons, which is 0.0058. 

A second hypothesis is that when $f=1$, the network has on average around $50\% \times Nhidden$ firing neurons for all of the images within the dataset. This is because $Sparsity \times Nin_b \times meanW$ represents the expectation of $x_n$ in Eq. (5), where $x_n=\sum_{m=0}^{M}x_m\times w_{mn}$. When $Sparsity \times Nin_b \times meanW=Vth$, a neuron in the hidden layer has on average 50\% possibilities to fire. When $f$ approaches zero, it is likely that none of the hidden layer neurons will fire, while when $f$ is much larger than one, it is likely that all of the hidden layer neurons will fire. The optimization goal is to keep $f$ close to one because the hybrid network has a greater learning capability when 50\% of the hidden layer neurons are firing. 

\begin{figure}
 \centering
 \vspace{-2em}
 \includegraphics[width=13.8cm,height=10cm]{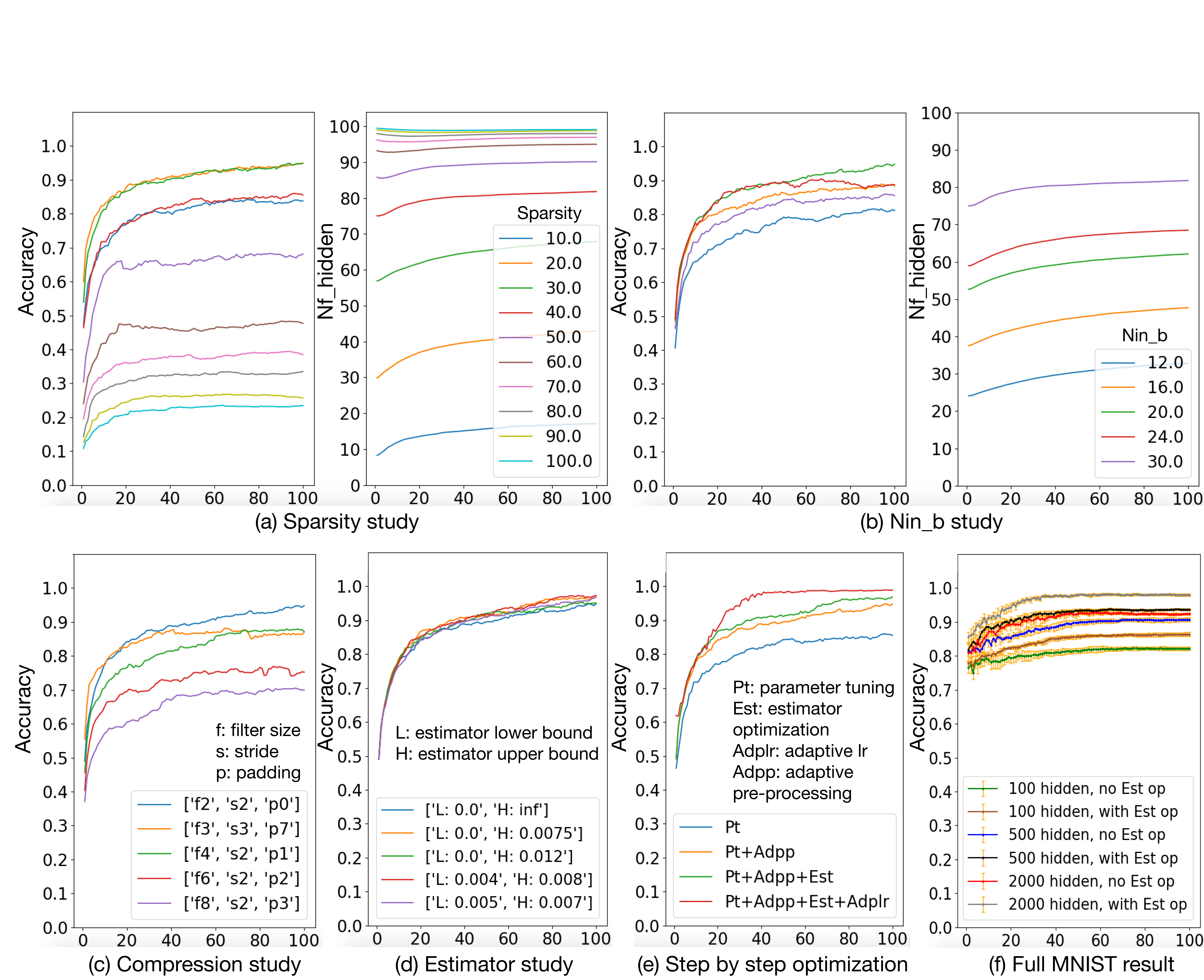}
 \vspace{-0.5em}
 \caption{Optimization study and the full MNIST result.}
 \vspace{-0.5em}
 \label{opt}
\end{figure}

An experiment is conducted to verify these two hypotheses by checking the influence of the network sparsity on the $Nf_{hidden}$ and the accuracy (Fig~\ref{opt} (a)). The results of ten different sparsity suggest that, when the sparsity is higher,  $Nf_{hidden}$ is larger. The best accuracy is achieved at the sparsity 20\% and 30\%, where the corresponding $Nf_{hidden}$s are closest to 50\% among the tested sparsities and the corresponding $f$ values are close to one. These observations agree with the hypotheses. 

\begin{table}
    \caption{Simulation parameters for the optimization study.}
    \label{1000setting}
    \footnotesize
    \centering
    \begin{tabular}{ll}
     \hline
     Data precision        &Floating point 64 for computational layer                \\
     Dataset/Network &1000 MNIST images, training set same as the testing set, size:196-100-10   \\ 
     Bio layer       &Sparsity:40\%, $Vth_{cp}$:N(0.0058, 0.0017), $initW$:N(0.0007, 0.0007), Lr:5e-6\\
     Hardware layer  &Fully connected, $initW$:N(0.0007, 0.03), Lr:0.008 \\
     Optimization    &Adpp:$Nin_b$=20, Estimator range:(0,0.0075) \\
                     &Adlr: bio layer initial Lr 5e-06, hardware layer initial Lr 0.1, decay rate 0.2 \\
     \hline
\end{tabular}
\vspace{-1.5em}
\end{table}
 
\subsection{Optimization Methods}
\vspace{-0.5em}
\textbf{Adaptive pre-procssing (Adpp)}. Based on the two hypotheses above, it is important to shift $Nf_{hidden}$ to 50\% for a living neural network with a certain sparsity. In this paper, an adaptive pre-procssing (Adpp) technique  for the living neural network is proposed. In this approach, the input images are processed to achieve a target $Nin_b$, so that the $f$ value is close to one. In Fig~\ref{opt} (b), sparsity 40\% is used as an example to test the Adpp approach. $Nin_b=30$ for the 1000 MNIST dataset before Adpp. To decrease the $Nf_{hidden}$ from 80\% to around 50\%, different $Nin_b$ are tested. Results show that, when $Nin_b=20$, $Nf_{hidden}$ is the closest to 50\%, and the corresponding accuracy is the highest (94.8\%). 

We adopt a filter-and-pool approach~\cite{cnn} as the pre-processing mechanism. It is sometimes necessary to compress the input data to fit the input-bio interface of the biological layer. Fortunately, the compression can be naturally incorporated into the pre-processing.
As described in Section~\ref{network}, the proposed work uses 196 neurons as the inputs. To compress the $28\times28$ MNIST images to $14\times14$, a specific filter size, stride, padding, and compression threshold need to be chosen. Relationship between these parameters are given by $compressed \, size=\frac{(original \, size-filter \, size+2\times padding)}{stride}+1$. 
As Fig~\ref{opt} (c) shows, different combinations of the filter size, stride, and padding are tested and the best results is given by filter size=2, stride=2, and padding=0. Therefore, theses set of pre-processing parameters are used in the experiment. The compression threshold that turns the grey scale value to black and white after the average pooling is also tuned  to meet the desired $Nin_b$ value. By using the Adpp approach, $Nf_{hidden}$ can usually be tuned to around 50\% for a typical living neural network sparsity, and thus a good accuracy can be achieved. The $f$ value is close to one when $Nf_{hidden}$ is close to 50\%. To find the best $Nin_b$ parameter for a specific experiment setup, a small dataset (100 MNIST data, 100 epoch) is used for $Nin_b$ tuning before performing the learning task in the target dataset. 

The average number of firing neurons can be used to explain the accuracy difference between the computational and the biophysical models (Fig~\ref{variation2}), which have 80\% and 60\% firing neurons in the hidden layer respectively at the 100 epoch. This explains the accuracy difference between these two models. The biophysical model with neuron variations decreased the number of firing neurons in the hidden layer, which improved the network accuracy. This result also suggests that the variation conversion method proposed in this paper tends to underestimate the variations in the computational model. However, the proposed Adpp method can still be used in the biophysical model to improve its accuracy.  

\textbf{Gradient estimator}. To further improve the hybrid network accuracy, this work studies different gradient estimators (Est). Because the hidden layer of the hybrid network uses a hard-threshold function, the gradient needs to be estimated. The straight through estimator~\cite{estimator}, which considers the gradient as constant one, is used for the previous experiments in this work. However, setting the gradient as one only when $x_n$ (Fig~\ref{algorithms}) is within a small range can improve the training of the network. A set of gradient ranges around Vth (0.0058) are tested. Results in Fig~\ref{opt} (d) show that passing the gradient across a neuron only when $x_n$ falls in a smaller range can help improve the accuracy. However, implementing this optimization approach on the living neural network is not easy. Because getting the accumulated current value (represented as $x_n$ in the biophysical model) instead of spike or not for each biological neuron requires intracellular recording on each hidden layer neurons. 
However, neuron fires earlier when the accumulated current is larger. By observing spikes earlier, it is possible to  cut off some weight updating when the accumulated current is greater than a value. To model this effect, estimators that only has an upper bound is explored, and the best accuracy is achieved  when the estimator range is (0,0.0075), this value is used for the following experiments.  

\textbf{Parameter tuning and adaptive learning rate}. The biological layer parameters such as the network sparsity, the initial weight distribution, and the threshold distributions use the fitted parameters from the realistic living neural network and thus cannot be changed. The biological layer learning rate and the computational layer parameters such as the initial weights, learning rate, and the connectivity can be tuned. The best parameters are found through an automatic searching of the parameter space, which are listed in Table \ref{1000setting}. A relatively large variance and learning rate for the computational layer can improve the convergence speed. The biological layer learning rate should be small because the weight is constrained within the 0.5--2$\times$ range of the initial weight. After parameter tuning, a 85.5\% testing accuracy is achieved at epoch 100 on the MNIST 1000 dataset (Fig~\ref{opt} (e)).
An adaptive learning rate (Adlr) method with exponential decay~\cite{binary} is applied to the network with parameters in Table~\ref{1000setting} and a 2\% accuracy improvement is achieved on the top of the previous optimization methods (Fig~\ref{opt} (e)).

\subsection{Results on the Full MNIST Dataset}
\vspace{-0.5em}
The hybrid bio-hardware neural network learning accuracy for the full MNIST dataset (training 60000, testing 10000) is reported in Fig~\ref{opt} (f) after adding all of the optimizations discussed above. The network parameters for this experiment are the same as the ones reported in Table~\ref{1000setting}, except for the optimized and adaptive parameters. The initial biological and hardware layer learning rate for full MNIST study is 1e-5 and 0.1 respectively, with a decay rate at 0.2. $Nin_b=26$. For 100, 500, and 2000 hidden layer neurons, the testing accuracy is 86.35 (+-0.63), 93.66 (+-0.3), 97.94 (+-0.37) respectively. The best accuracy achieves at 98.3\% with 2000 hidden layer neurons, which is similar to the reported MNIST accuracy (99.04\%) of a three-layer binary network with 4096 binary unit in each layer~\cite{binary}. 

\vspace{-0.5em}
\section{Conclusion}
\vspace{-0.5em}
In this paper, a hybrid bio-hardware neural network is proposed and studied using both biophysical and computational models. The biological layer faithfully models the living neural network properties such as the neuron and synapse variations, synapse constraints, and network connectivity. 
Several techniques are proposed
to improve the inference accuracy of the proposed hybrid neural network. 
A near the state-of-the-art accuracy is achieved in simulation using a living neural network fitted computational model. Our work demonstrates the feasibility of using living neural networks with high levels of inherent variability to perform machine learning tasks and achieve good inference accuracy. 

\begin{spacing}{0.83}
{
\small
\bibliographystyle{plain}
\bibliography{refs}
}
\end{spacing} 

\newpage
\section*{Supplementary Materials}

\subsection*{Biophysical Equations and Terminologies}
The Pinsky-Renzel ~\cite{pr2} neuron model is described by Eq. (1) and (2). The alpha synapse ~\cite{alpha} model is described by Eq. (3) and (4). The neuroscience terminologies used in this paper are summarized in Table~\ref{term}.

\vspace{-1em}
\begin{equation}
\begin{aligned}
C_m \frac{dV_s}{dt}=&-\bar{g}_L(V_s-E_L)-g_{Na}(V_s-E_{Na})
\\&-g_{DR}(V_s-E_k) +\frac{g_c}{p}(V_d-V_s)+ \frac{I_s}{p} 
\end{aligned}
\end{equation}
\vspace{-2.5em}

\begin{equation}
\begin{aligned}
C_m \frac{dV_d}{dt}=&-\bar{g}_L(V_d-E_L)-g_{Ca}(V_d-E_{Ca})
\\&-g_{AHP}(V_d-E_k)+\frac{g_c}{1-p}(V_s-V_d)+\frac{I_{syn}}{1-p}
\end{aligned}
\end{equation}
\vspace{-2.5em}

\begin{equation}
I_{syn}(t)=g_{syn}(t)(V_d(t)-E_{syn}) 
\end{equation}
\vspace{-2.5em}

\begin{equation}
g_{syn}(t)=\bar{g}_{syn} \frac{t-t_s}{\tau} exp(-\frac{t-t_s}{\tau}) 
\end{equation}
\vspace{-1.5em}

\begin{table}[bp]
    \footnotesize
    \caption{Terms used in this paper~\cite{pr}~\cite{pr2}~\cite{alpha}.}
    \label{term}
    \centering
    \begin{tabular}{ll}
     \hline
     $C_m$                                    & membrane capacitance\\
     $E_{L}$                                  & membrane resting potential\\
     $g_c$                                    & soma–dendrite coupling conductance\\
     $p$                                      & the proportion of the membrane area occupied by soma \\
     $\bar{g}_{L}$                            & leakage conductance\\
     $\bar{g}_{Na}$                           & maximum conductance of sodium current\\
     $\bar{g}_{DR}$                           & maximum conductance of outward delayed-rectifier potassium current\\
     $\bar{g}_{AHP}$                          & maximum conductance of potassium after hyper-polarization current \\
     $\bar{g}_{Ca}$                           & maximum conductance of calcium current\\
     $\bar{g}_{C}$                            & maxium conductance of calcium-dependent potassium current \\
     $\bar{g}_{syn}$                          & maximum synaptic conductance\\
     $\tau$                                   & time constant of synaptic conductance \\
     $E_{syn}$                                  & reversal potential of the synaptic ion channels \\
     $t_s$                                    & time when a pre-synaptic spike arrives\\
     $synapse$ $transmutation$ $delay$        & time between the pre-synaptic peak potential and the point\\
                                              & when post-synaptic potential starts to change \\
     $Time$ $to$ $first$ $spike$              & time between the current injection and the maximal slope of\\
                                              & the post synaptic membrane potential\\
     $After$-$hyperpolarization$ $voltage$    & The minimum voltage after the first spike and before the \\
                                              & following spikes (if have) \\
     \hline
     
    \hline
    \end{tabular}
\end{table}

\subsection*{Parameter Fitting}
Neural cultures were obtained by dissociating cortices of postnatal day 0 Sprague Dawley rats and plating neurons onto poly-D-lysine coated tissue culture dishes.  
On days \textit{in vitro} (DIV) 12 - 19, neuron IV characteristics were obtained by injecting currents from -200 to 300 pA in current clamp mode with 25 pA delta current step.
Biophysical neural model (Pinsky-Rinzel) parameters were then adjusted until they produced similar spiking behavior (time to first spike, number of spikes, and after-hyperpolarization voltage produced by current injection) as nine recorded cells.   Excitatory postsynaptic currents (EPSCs) were evoked by patterned blue light stimulation of ChR2-expressing pre-synaptic neurons.  Synaptic parameters were then extracted by fitting an alpha function to experimentally obtained EPSC waveforms for 12 different post-synaptic neurons.  Experimental data represent an average of 10 trials per each cell.  Parameters of the experimentally-matched biophysically modelled cells and synapses are listed in Table 4.  All animal experimental protocols were approved by the Institution Animal Care and Use Committee (IACUC) at Lehigh University and were conducted in accordance with the United States Public Health Service Policy on Humane Care and Use of Laboratory Animals.

\begin{table}[htbp]
  \centering
  \footnotesize
  \caption{Parameters Fitting Results.}
  \vspace{-0.5em}
    \begin{tabular}{lcccccccccccccc}
    \hline
    \multicolumn{11}{c}{\textbf{Neuron Parameters}}\\
    \hline
    Cell  & 1     & 2     & 3     & 4     & 5     & 6     & 7     & 8     & 9     & Mean \\
    \hline
    $c_m (\mu F/cm^2)$   & 10    & 8     & 10    & 15    & 15    & 10    & 8     & 10    & 8     & 10.44 \\ 
    $E_{L} (mV)$ & -45   & -40   & -45   & -35   & -45   & -56   & -50   & -60   & -60   & -48.44 \\ 
    $\bar{g}_{L}\times 10^{-3} (S/cm^2)$  & 1.10   & 0.85  & 1.48  & 1.15  & 1.48  & 2.10   & 0.8   & 1.10   & 0.70   & 1.20  \\ 
    $\bar{g}_{Na}\times 10^{-2} (S/cm^2)$  & 8     & 6     & 15    & 29    & 25    & 9     & 7     & 11    & 8     & 13.11 \\
    $\bar{g}_{DR}\times 10^{-2}(S/cm^2)$  &2     & 2     & 0.5   & 2     & 1     & 9.9   & 2     & 12    & 6     & 4.16  \\
    $\bar{g}_{AHP}\times 10^{-2}(S/cm^2)$  & 1     & 0.9   & 0.1   & 0.45  & 0.1   & 3     & 0.5   & 0.5   & 12    & 2.06  \\
    $\bar{g}_{Ca}\times 10^{-2}(S/cm^2)$  & 0.8   & 0.8   & 0.8   & 0.8   & 0.8   & 0.8   & 0.8   & 0.8   & 0.8   & 0.80  \\
    $\bar{g}_{C}\times 10^{-2}(S/cm^2)$  & 2     & 2     & 2     & 2.5   & 10    & 20    & 2     & 2     & 10    & 5.83  \\
    \hline
    \multicolumn{11}{l}{*cell diameter is 20$um$, $p$ is 0.5, and $g_c$ is 8 $(S/cm^2)$ for all cells}\\
    \hline
    \end{tabular}%
  
    \begin{tabular}{lccccccccccccc}
    \multicolumn{14}{c}{\textbf{Synapse Parameters}}\\
    \hline
    Cell   & 1     & 2     & 3     & 4     & 5     & 6     & 7     & 8     & 9     & 10    & 11    & 12    & Mean  \\
    \hline
    $gsyn$ & 0.72  & 0.59  & 0.34  & 0.73  & 0.69  & 0.63  & 1.15  & 0.21  & 1.17  & 2.28  & 0.59  & 0.41  & 0.79  \\
    $\tau$   & 4.50  & 5.70  & 4.55  & 7.80  & 5.75  & 6.96  & 5.95  & 6.00  & 4.40  & 4.75  & 4.90  & 6.99  & 5.69  \\ 
    $delay$ & 1.80  & 2.00  & 2.00  & 2.00  & 2.00  & 1.30  & 2.00  & 2.00  & 1.40  & 1.30  & 2.00  & 2.00  & 1.82  \\
    \hline
    \multicolumn{14}{l}{*the unit for $gsyn$ is $\times 10^{-3} (\mu S)$, the unit for delay is $(ms)$} \\
    \hline
    \end{tabular}%
    
  \label{fitting_result}%
\end{table}%


\subsection*{A Study of the Random Network}
\label{recurrent}

\begin{figure}[bp]
  \centering
  \includegraphics[width=13.8cm,height=7cm]{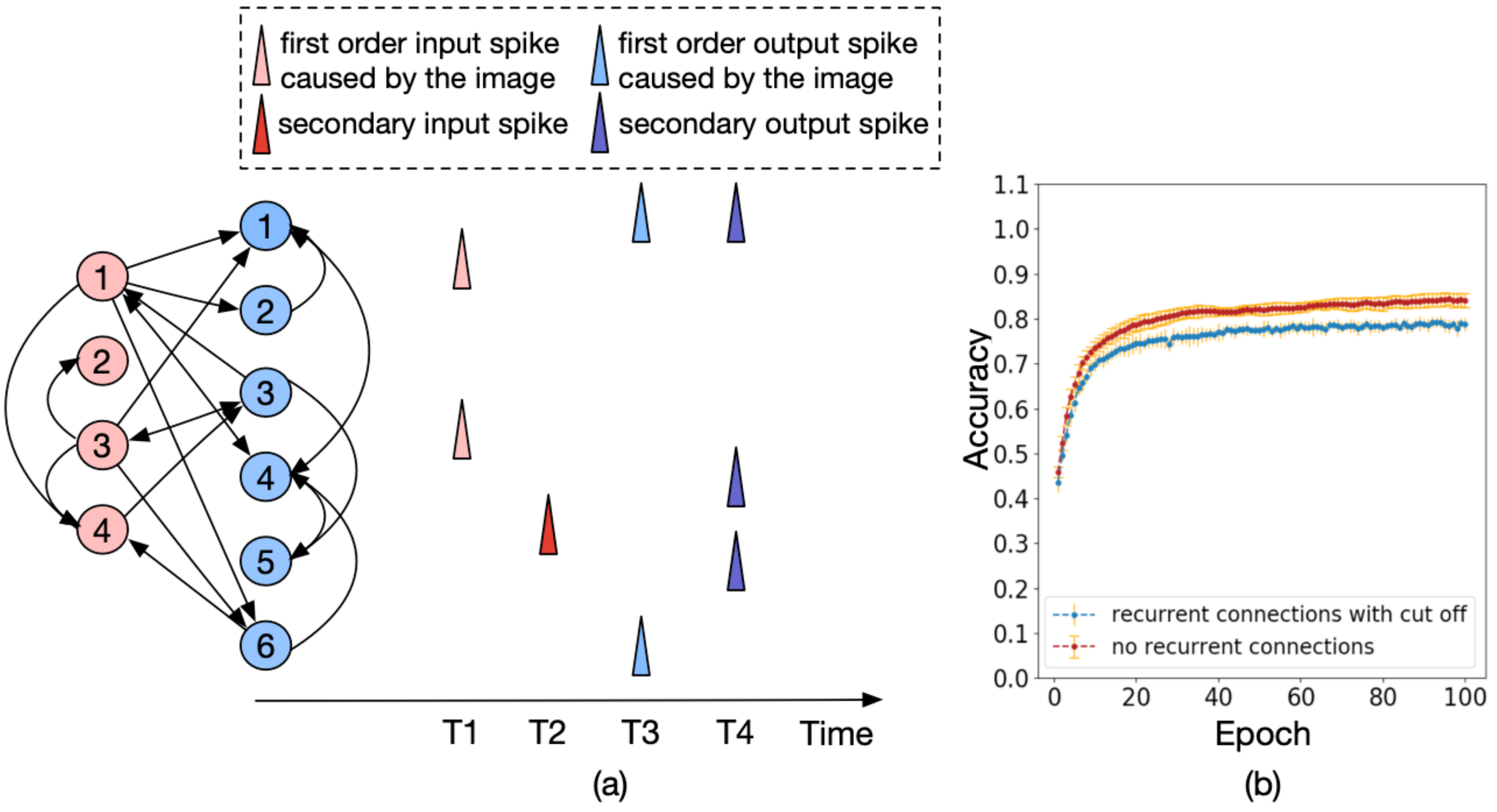}
  \caption{(a) An example of the input and output spike pattern in a random connection network (b) An example of the output spike pattern given by the NEURON simulator.}
  \label{recurrent1}
\end{figure}

\begin{figure}
  \centering
  \includegraphics[width=13.8cm,height=5cm]{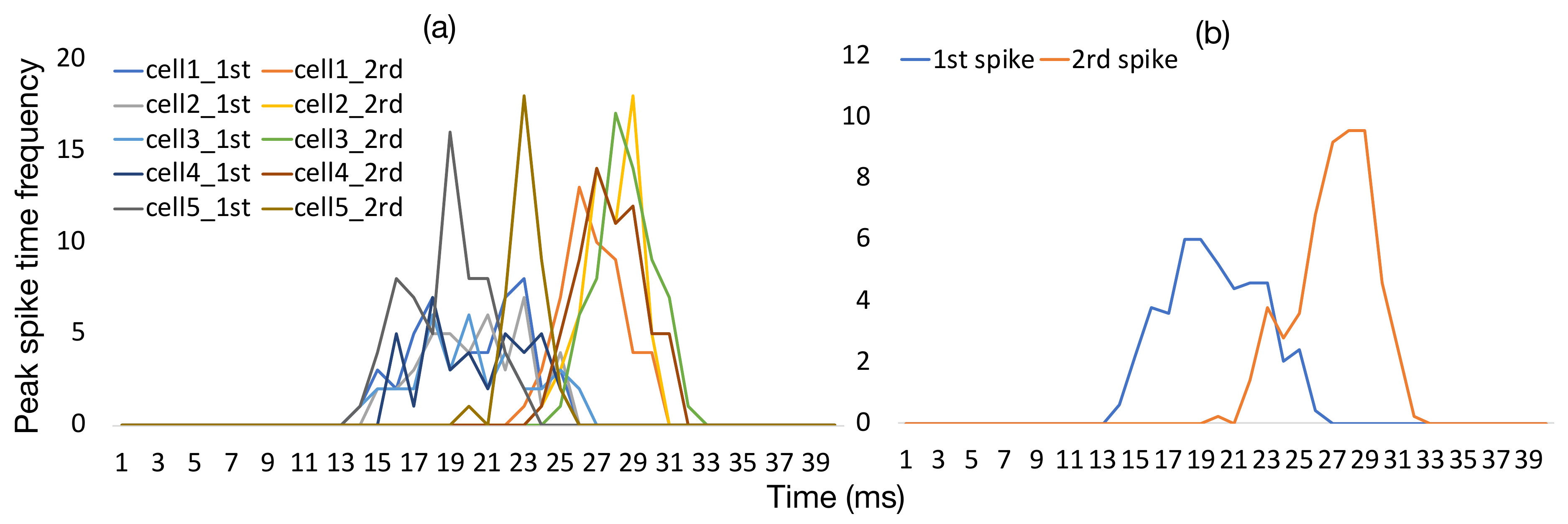}
  \caption{A study of the primary and secondary spikes.}
  \label{recurrent2}
\end{figure}

The living neural network is randomly connected, which means that a connection is possible between any two neurons.
A challenge due to random connections is how to differentiate the secondary spikes from the primary spikes (Section \ref{method}). An example of the input and output spike pattern is given in Fig~\ref{recurrent1} (a). In this example, it is assumed that two input spikes will cause an output spike. When an image is given to the network at time T1, it generates two primary output spikes for output neuron 1 and 6 at time T3. However, since the network has random connections, unexpected secondary spikes at input 4, output 1, output4, and output 5 are incurred.  In this example, the spike generation delay (time between the input spike peak and the output spike peak) are assumed to be the same, so that the target output results can be recognized at T3. However, when the neuron dynamics and variations are modeled, the target primary output spikes do not happen at the same time. Furthermore, neuroscience studies indicate that the spike generation delay is in direct proportion to the neuron distance~\cite{scaling}, which means that some primary spikes can happen after some secondary spikes, which can cause issues during both inference and training.

To examine how many of the primary and the secondary spikes overlap, an experiment is conducted in the NEURON simulator and the results are shown in Fig~\ref{recurrent2}. A 40\% randomly connected 196-100 feedforward network (only input-to-output connections are considered) with the neuron, the synapse, and the spike generation delay variations is built. An image is given to the network and the output spike pattern is recorded as the primary spikes. Then random connections are added to the same network, with the same random initialization. When given the same image, the new output spike pattern is recorded. The difference between the new and old spike patterns in the two recordings are the secondary spikes. This experiment is conducted for five times with different MNIST images. The peak spike time is recorded for each output neuron if a spike is generated and the frequency of the peak spike time is reported in Fig~\ref{recurrent2} (a). Fig~\ref{recurrent2} (b) is the average frequency of peak spike time among five cells, and shows that only a small portion (roughly 10\%) of the primary and secondary spikes overlap. 

To minimize the influence of the random connections on the prediction accuracy, an early cut off mechanism is proposed, which consider only the spikes that happen before a certain cutoff time as the output of the network. In the Fig~\ref{recurrent2} example, when the cut off time is chosen at 24ms, roughly 10\% of the primary spikes are lost and 10\% of the secondary spikes are considered as the network output. Figure~\ref{recurrent1} (b) shows the accuracy influence of the cut off scheme in the computational model and an average 6\% accuracy drop is incurred among 10 runs. Experimental configurations are the same as Table~\ref{100setting} without applying any of the optimization techniques.

\end{document}